\documentclass[runningheads]{llncs}
\usepackage{microtype}
\usepackage{graphicx}
\usepackage{tikz}
\usetikzlibrary{
    backgrounds,
    calc,
    positioning,
    scopes,
    shadings,
    shadows,
}
\usepackage{bm}
\usepackage{subfigure}
\usepackage{nicefrac}       
\usepackage{siunitx}
\usepackage[utf8]{inputenc} 
\usepackage[T1]{fontenc}    
\usepackage{booktabs} 
\usepackage{environ}
\usepackage{tabularx}
\usepackage{multirow}
\usepackage{makecell}
\usepackage{xfrac}
\usepackage[misc]{ifsym}
\usepackage{breqn} 
\usepackage{graphicx}

\usepackage{biblatex}[style=splncs04,sorting=none]
\usepackage{hyperref}

\usepackage{color}

\usepackage[capitalize,noabbrev]{cleveref}

\newcommand\mvec[1]{\bm{#1}}

\newcommand\tnstd{standard}
\newcommand\tnres{additive}
\newcommand\tnmod{multiplicative}
\newcommand\tnshd{shared}

\makeatletter
\newcommand{\printfnsymbol}[1]{%
  \textsuperscript{\@fnsymbol{#1}}%
}
\makeatother

\begin{document}
\title{Hypernetworks build Implicit Neural Representations of Sounds}

\author{
Filip Szatkowski\thanks{equal contribution}\inst{1,3}\orcidID{0000-0001-8592-2001}{ \Letter}
\and\\
Karol J. Piczak\printfnsymbol{1}\inst{2}\orcidID{0000-0002-6115-0833}
\and\\
Przemysław Spurek\inst{2}\orcidID{0000-0003-0097-5521}
\and\\
Jacek Tabor\inst{2,5}\orcidID{0000-0001-6652-7727}
\and\\
Tomasz Trzciński\inst{1,2,3,4}\orcidID{0000-0002-1486-8906}
}
\authorrunning{Szatkowski et al.}

\toctitle{Hypernetworks build Implicit Neural Representations of Sounds}
\tocauthor{Filip~Szatkowski, Karol~J.~Piczak, Przeymysław~Spurek, Jacek~Tabor, Tomasz~Trzciński}

\institute{
Warsaw University of Technology, Poland\\\email{\{filip.szatkowski.dokt,tomasz.trzcinski\}@pw.edu.pl}
\and
Faculty of Mathematics and Computer Science, Jagiellonian University, Poland\\\email{\{karol.piczak,przemyslaw.spurek,jacek.tabor\}@uj.edu.pl}
\and
IDEAS NCBR
\and
Tooploox
\and
UES Ltd.
}


%
\maketitle              
\begin{abstract}
Implicit Neural Representations (INRs) are nowadays used to represent multimedia signals across various real-life applications, including image super-resolution, image compression, or 3D rendering. Existing methods that leverage INRs are predominantly focused on visual data, as their application to other modalities, such as audio, is nontrivial due to the inductive biases present in architectural attributes of image-based INR models. To address this limitation, we introduce HyperSound, the first meta-learning approach to produce INRs for audio samples that leverages hypernetworks to generalize beyond samples observed in training. Our approach reconstructs audio samples with quality comparable to other state-of-the-art models and provides a viable alternative to contemporary sound representations used in deep neural networks for audio processing, such as spectrograms. Our code is publicly available at \href{https://github.com/WUT-AI/hypersound}{https://github.com/WUT-AI/hypersound}.

\end{abstract}

\section{Introduction}
\begin{figure}[!ht]
    \centering
    \includegraphics[width=0.8\columnwidth,trim={0cm 0.5cm 0cm 0cm},clip]{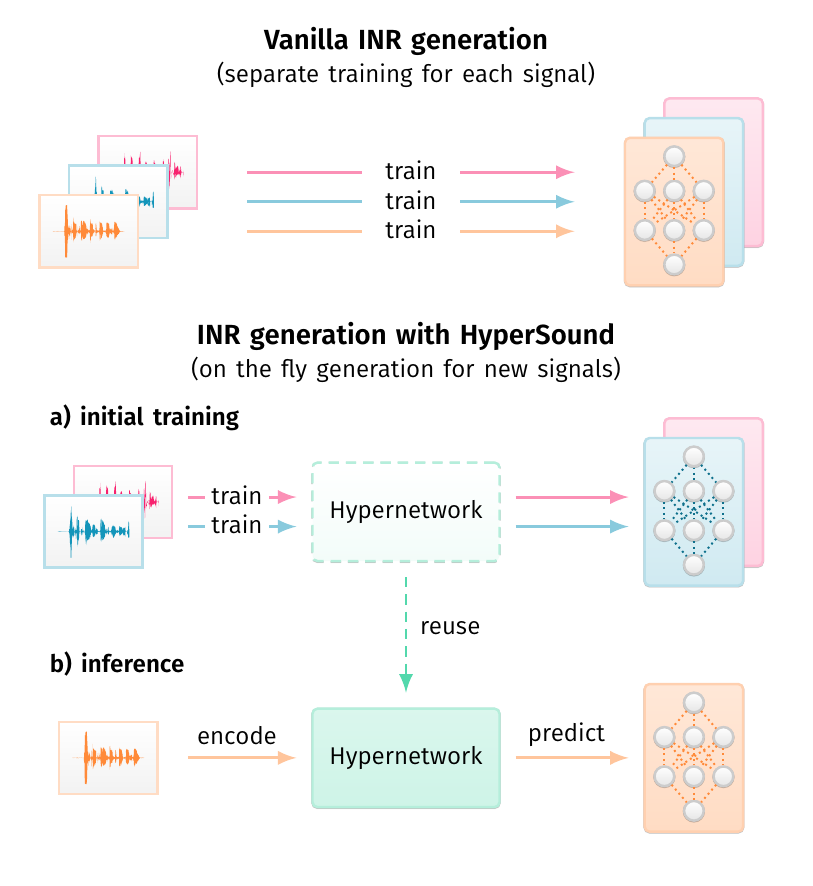}
    \vspace{-1em}
    \caption{We propose HyperSound, the first framework which produces Implicit Neural Representations for unseen audio signals, eliminating the need for retraining a model for each new example.} 
\end{figure}

The field of Implicit Neural Representations (INRs) is a rapidly growing area of research, which aims to obtain functional, coordinate-based representations of multimedia signals through the use of neural networks. These representations are decoupled from the spatial resolution, so the signal can be resampled at any arbitrary frequency without incurring additional memory storage requirements. Applications of INRs include super-resolution~\cite{sitzmann2020implicit,mehta2021modulated}, compression~\cite{sitzmann2020implicit,mehta2021modulated}, 3D rendering~\cite{mildenhall2021nerf}, and missing data imputation~\cite{fons2022hypertime}. Furthermore, INRs can be used to synthesise data~\cite{zuiderveld2021towards}, and they show promise as an alternative to the traditional feature extraction techniques used in neural networks~\cite{dupont2022data}.

INRs are a promising research direction in audio processing, yet so far their applications in this domain remain limited. As the size of INR networks can be significantly smaller than the size of the represented signal, they can be used as a compression method~\cite{strumpler2022implicit}. They also provide an alternative to super-resolution algorithms such as ProGAN~\cite{karras2018progressive}, which often require hand-crafting complex architectures for selected resolution. In contrast, in INRs the signal resolution is given by the granularity of input coordinates, so a single network represents the signal for any resolution we might require while remaining constant in size. Moreover, as ~\cite{muller2022diffrf} show that INRs can be used as features in other algorithms, this representation can be an alternative for current deep audio processing models that primarily operate on features obtained from mel spectrograms or raw audio waves~\cite{purwins2019deep}. 

In practice, architectures typically used for INRs struggle with modeling high frequencies inherent in audio signals~\cite{rahaman2019spectral} and require training a new model for every incoming data point, which is wasteful and greatly reduces their applicability. This issue is usually addressed by obtaining generalizable INRs through meta-learning methods, such as hypernetworks~\cite{ha2016hypernetworks} -- models that produce the weights for other models. The hypernetwork framework can be leveraged to obtain generalizable INRs for a variety of data types, including images~\cite{sitzmann2020implicit,mehta2021modulated,klocek2019hypernetwork}, point clouds~\cite{mehta2021modulated,spurek2020hypernetwork,spurek2022hyperpocket} 3D objects~\cite{kania2023hypernerfgan,zimny2022points2nerf}, videos~\cite{mehta2021modulated} and time series~\cite{fons2022hypertime}. However, obtaining generalizable INRs for audio signals remains an unsolved task due to the high dimensionality and variance of this type of data, which makes the standard methods for this task insufficient.

In our work, we propose a method that learns a general recipe for creating INRs for arbitrary audio samples not present in the training dataset. We adapt the hypernetwork framework for sound processing by introducing a domain-specific loss function and hypernetwork weight initialization. Additionally, we investigate various variants of the hypernetwork framework in order to achieve improved compression while maintaining audio quality. To our knowledge, our method, named HyperSound, is the first application of hypernetwork-based INR generation to the audio domain.

The main contributions of our work are:
\begin{itemize}
    \item We introduce HyperSound, to our knowledge the first INR method for audio processing that builds functional representations of unseen audio signals, which eliminates the need of retraining the INR for each new data point.
    \item We show a successful adaptation of the hypernetwork paradigm to the audio processing through domain-specific modifications such as the spectral loss function.
    \item We empirically evaluate various approaches of using hypernetworks to build INRs for audio generation, e.g. by leveraging SIREN architecture or Fourier mapping.
\end{itemize}

\section{Related Works}

\subsection{Implicit Neural Representations (INRs)}

Implicit Neural Representations (INRs) are continuous, coordinate-based approximations of data obtained through neural networks. The decoupling of representation from spatial resolution makes INRs attractive for a wide range of applications in signal processing. However, functions represented by standard MLP networks with ReLU activation function show a strong bias towards low frequencies~\cite{rahaman2019spectral}, which in practice make them infeasible as representations of multimedia signals. To mitigate this bias, most INRs use either Fourier features~\cite{tancik2020fourier} or SIREN networks~\cite{sitzmann2020implicit} with periodic activation functions. Later works improve SIREN networks, introducing various forms of modulation to sinusoidal layers~\cite{mehta2021modulated,chan2021pi}. INRs are currently used in many applications, such as super-resolution~\cite{sitzmann2020implicit,mehta2021modulated}, compression~\cite{strumpler2022implicit}, 3D rendering~\cite{mildenhall2021nerf}, missing data imputation~\cite{fons2022hypertime} or data generation~\cite{zuiderveld2021towards}. However, typical INRs are trained for each data point, and obtaining a representation for new signals requires retraining the network from scratch.

\subsection{Generalizable INRs}
To obtain a way of generating INRs for new data without the need for model retraining, various meta-learning approaches were proposed. \cite{sitzmann2020metasdf} leverage gradient-based meta-learning algorithms to solve the task of learning an INR space for shapes. \cite{chen2022transformers} employ a~transformer meta-learner to obtain INRs for unseen examples. Many methods of obtaining generalizable INRs also involve the use of hypernetworks~\cite{ha2016hypernetworks}, where one network (hypernetwork) generates the weights for another network (target network). More broadly, hypernetworks are employed in many applications, including model compression~\cite{zhao2020meta}, continual learning~\cite{von2019continual} or generating INRs. In particular, hypernetworks were successfully used to generate INRs for images~\cite{klocek2019hypernetwork,sitzmann2020implicit,mehta2021modulated}, shapes~\cite{spurek2020hypernetwork,sitzmann2020implicit}, videos~\cite{mehta2021modulated} and time series~\cite{fons2022hypertime}. To our knowledge, our work is the first application of hypernetworks for generating audio INRs.

\subsection{Deep neural networks for audio processing}
First successful attempts at raw waveform processing with deep neural networks were models such as WaveNet~\cite{oord2016wavenet} and SampleRNN~\cite{mehri2016samplernn}, but their autoregressive nature makes them slow and prone to accumulation of errors. Later architectures such as ParallelWaveNet~\cite{oord2018parallel}, NSynth~\cite{engel2017neural}, MelGAN~\cite{kumar2019melgan} or SING~\cite{defossez2018sing} proposed non-autoregressive architectures for audio generation. Recent autoencoder-based models such as RAVE~\cite{caillon2021rave} or SoundStream~\cite{zeghidour2021soundstream} are able to process high-resolution signals in an end-to-end fashion, producing audio of very good perceptual quality.

\section{Method}
\label{sec:method}

\begin{figure*}[!t]
    \centering
    \includegraphics[width=\textwidth]{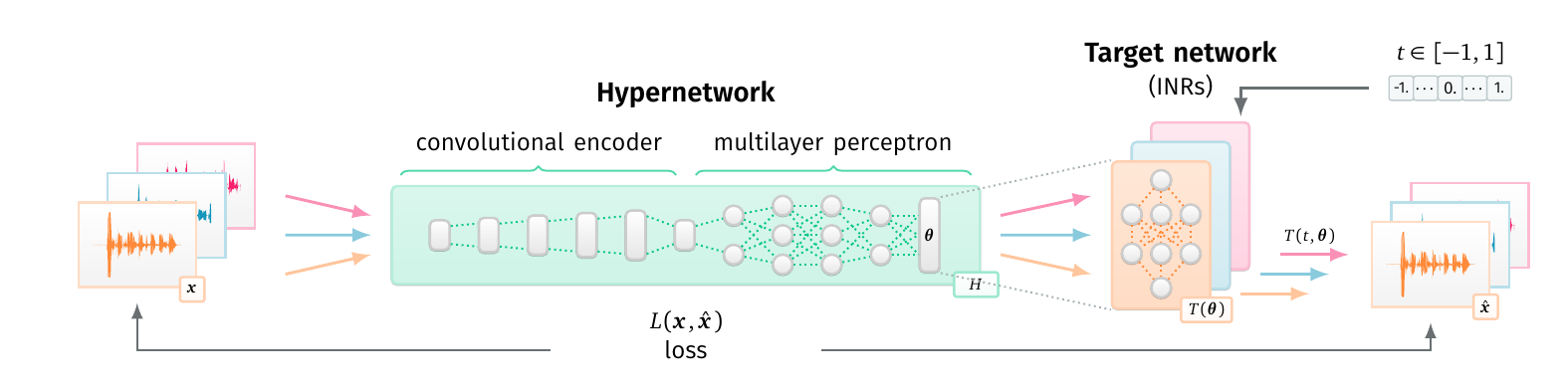}
    \vspace{-1.5em}
    \caption{Overview of the HyperSound framework. We use a~single hypernetwork model to produce distinct INRs based on arbitrary audio signals provided as input.} 
    \label{fig:teaser}
\end{figure*}

Traditional digital representation of sound waves is a series of amplitude values sampled at regular time intervals, which approximates a~continuous real function $x(t)$. Our goal is to obtain a~meta-recipe for generating audio INRs approximating such functions. While the creation of INRs for particular audio samples can be successfully achieved with gradient descent, finding a general solution is much harder due to the inherent complexity of audio time series. Therefore, similar to \cite{klocek2019hypernetwork}, we model such functions with neural networks $T$ (target networks), parameterized by weights generated by another neural network $H$ (hypernetwork). Our method, named HyperSound, can be described as
\begin{equation}
\begin{gathered}
    \mvec{\theta}_{\mvec{x}} = H(\mvec{x}),\\
    \hat{x}(t) = T(t, {\mvec{\theta}}_{\mvec{x}}).
\end{gathered}
\end{equation}

We show the overview of our method in Fig.~\ref{fig:teaser}. In the following sections, we describe the architecture of the models used as Implicit Neural Representations and of the hypernetwork. Then, we present the details required for the successful application of the hypernetwork framework to the audio domain.

\subsection{Implicit Neural Representations}
\label{sec:method:inr}

We consider two variants of INR networks, inspired by NeRF~\cite{mildenhall2021nerf} and SIREN~\cite{sitzmann2020implicit} papers. Both models have a single input and a single output. 

We refer to a NeRF-inspired network as a~Fourier-mapped multilayer perceptron (FMLP). The network consists of a~positional encoding layer followed by dense layers with biases and ReLU activation functions. Positional embedding vectors $\mvec{\gamma}$ are obtained as

\begin{equation}
    \mvec{\gamma}(t) =\Bigl[\sin(2^0 \pi t), \cos(2^0 \pi t), \ldots, \sin({2}^{L-1} \pi t), \cos({2}^{L-1} \pi t)\Bigr]
\end{equation}
where $t$ is the time coordinate and $L$ denotes the embedding size. We normalize input coordinates to a~range of $[-1, 1]$ and use $L=10$.

SIREN network consists of dense layers with a sinusoidal activation function. The output $\mvec{y_{i}}$ of $i$-th SIREN layer is defined as:
\begin{equation}
    \mvec{y_{i}} = \sin{(\omega_{i} \mvec{x_{i}} \mvec{W_{i}}^T + \mvec{b_{i}})},
\end{equation}
where $x_{i}$ is the input of the layer, $\mvec{W_{i}}$ and $\mvec{b_{i}}$ are layer weights and biases and $\omega_{i}$ is a scalar factor introduced in \cite{sitzmann2020implicit} to enable learning of higher signal frequencies. In practice, following the original paper, we set a distinct $\omega_{0} \neq 1$ for the first layer and keep the same shared value of $\omega_{i}$ for all remaining layers.

\cite{benbarka2022seeing} prove that FMLP networks are equal to single-layer SIREN networks. However, in our experiments, we see significant differences in the learning dynamics and representational capacity of both networks and thus explore both architectures in our experiments.

\subsection{Hypernetwork}
\label{sec:method:hn}

Typical audio recordings contain several thousands of samples, while hypernetworks generally use fully connected layers to produce target weights. Since we want to use raw audio recordings as input, using dense layers to process the signal directly leads to the explosion of the number of weights in the hypernetwork. To avoid this problem, we use a convolutional encoder to produce a~latent, lower dimensional representation of the signal and then process this representation with fully connected layers. The hypernetwork outputs weights $\mvec{\theta}$ that parameterise the target network.

We use an encoder based on SoundStream~\cite{zeghidour2021soundstream}. The fully connected part of the hypernetwork is composed of six layers with biases and ELU~\cite{clevert2015fast} activation, where the last layer produces the flattened weights of the target network. Input of the hypernetwork is  \num{32768} samples of audio at a sampling rate of~\num{22050}~Hz, the size of a~latent representation is $103 \times 32$ and the layer sizes of the dense part are \num{400}, \num{768}, \num{768}, \num{768}, \num{768}, \num{768}, \num{400}. 

\subsection{Hypernetwork adaptation for audio data}
\label{sec:method:hn_init}
Audio signals contain inherent high-frequency content, and neural networks in general tend to focus on low-frequency features~\cite{rahaman2019spectral}. Using a hypernetwork to produce INR weights introduces additional complexity to the optimization process. During training, small updates to the hypernetwork weights might disproportionally affect the function represented by the output weights, which makes the training process unstable. Moreover, the outputs of the neural network will always contain some inherent noise, which affects the high-frequency content more. Finally, in the case of training INRs, hypernetworks tend to overfit the training data, as they usually require over-parametrization to be able to solve a difficult problem of function generation. Combined, those issues make the straightforward application of the hypernetwork framework to our problem impossible.

To enable hypernetworks to model audio data, we introduce STFT loss function, which counters the hypernetworks tendency to ignore high-frequencies. The optimization process is described in detail in Section~\ref{sec:method:optim}. The spectral loss component is crucial to obtain the INRs of sufficient quality, as using only standard reconstruction loss in the time domain leads to the network collapsing to trivial solutions such as uniform noise or silence. Additionally, we notice that our method tends to overfit the training dataset and find that in order to make hypernetwork generalize to the unseen data we have to introduce random data augmentations to the training process. 

\subsection{Optimization for audio reconstruction}
\label{sec:method:optim}

We train the hypernetwork in a supervised fashion using backpropagation. To obtain audio results that are more perceptually pleasant, we use a~loss function that penalizes the reconstruction error both in time and frequency domains. Given an original recording $\mvec{x}$ and its reconstruction $\hat{\mvec{x}}$ generated with a target network, we compute the loss function as:

\begin{dmath}
    \label{eq_loss}
    L(\mvec{x}, \hat{\mvec{x}}) = {\lambda}_{t} L_{t}(\mvec{x}, \hat{\mvec{x}}) + {\lambda}_{f} L_{f}(\mvec{x}, \hat{\mvec{x}}),
\end{dmath}
where $ L_{t}$ is the L1 loss in time domain and $L_{f}$ is a~multi-resolution mel-scale STFT loss in frequency domain as introduced in ParallelWaveGAN~\cite{yamamoto2020parallel}. We use coefficients ${\lambda}_{t}$ and ${\lambda}_{f}$ to control the tradeoff between both losses.

For the STFT loss, we use \num{128} mel bins and FFT sizes of \num{2048}, \num{1024}, \num{512}, \num{256}, \num{128} with matching window sizes and an overlap of \num{75}\%. We set ${\lambda}_{t}=1$ and ${\lambda}_{f}=1$. Using both loss functions together allows us to optimize both quantitative metrics through L1 part of the loss and the perceptual quality of generated audio through the STFT part. We base this intuition on results from \cite{fons2022hypertime}, which show that the FFT loss is crucial for learning INRs for time series.

Neural networks are naturally biased towards low-frequency features, and we notice that the target networks obtained with HyperSound suffer from the same problem. Therefore, in some experiments we modify the STFT loss by frequency-weighting the Fourier transforms of $\mvec{x}$ and $\mvec{\hat{x}}$. We compute the weightings $w_{f_{i}}$ for frequency bins $f_{1}, f_{2}, ..., f_{N}$ as:
\begin{equation}
    w_{f_{i}} = \frac{N \cdot i^{p}}{\sum_{j=0}^{N}j^{p}},
\end{equation}
where $N$ is the number of frequency bins and $p\geq0$ is a parameter that allows us to control the uniformity of this weighting. The weights are normalized so that they always sum to $N$, and the minimum value of $p$ yields a uniform weighting (no bias towards high frequencies). During training, we gradually move from uniform weighting to target weighting in 500 epochs.

\subsection{Alternative ways to leverage hypernetworks for INR generation}

\label{sec:method:tn_types}

\begin{table*}[!t]
\caption{Comparison of different variants of computing target network pre-activation layer output.}
\label{tab:tn_types}
\begin{tabular*}{\textwidth}{lXlll}
\toprule

\multirow{3}{2cm}{\textbf{Network type\hspace{0.5cm}}} & \multirow{3}{2cm}{\textbf{Instance-specific weights}} & \multirow{3}{2cm}{\textbf{Shared weights}} & \multirow{3}{*}{\textbf{Pre-activation layer output}} \\ \\ \\

\midrule

    \tnstd            & all layers        & none          & $\mvec{y_{i}} = \mvec{x}(\mvec{W_{i}^{h}})^T + \mvec{b_{i}^{h}}$             \\[0.95em]
    \tnres            & all layers        & all layers    & $\mvec{y_{i}} = \mvec{x}(\mvec{W_{i}^{s}} + \mvec{W_{i}^{h}})^T + \mvec{b_{i}^{s}} + \mvec{b_{i}^{h}}$   \\[0.95em]
    \tnmod           & all layers        & all layers    & $\mvec{y_{i}} = \mvec{x}(\mvec{W_{i}^{s}} \cdot \mvec{W_{i}^{h}})^T + \mvec{b_{i}^{s}} \cdot \mvec{b_{i}^{h}}$   \\[0.95em]
    \tnshd    & some layers       & some layers   & $\mvec{y_{i}} = \begin{cases}
    \mvec{x}(\mvec{W_{i}^{s}})^T + \mvec{b_{i}^{s}}, & \text{if layer shared}\\
    \mvec{x}(\mvec{W_{i}^{h}})^T + \mvec{b_{i}^{h}},              & \text{otherwise}
\end{cases}$ \\

\bottomrule

\end{tabular*}
\end{table*}

In the \tnstd{} variant of our framework, the hypernetwork generates weights and biases for target network layers and other parameters, such as frequencies of positional encoding in FMLP or $\mvec{\omega_{i}}$ in SIREN, remain fixed. However, we also consider learning a shared, generic set of weights and biases for target network layers and combining them with instance-specific weights and biases generated by the hypernetwork in other ways. We refer to the three variants of target networks explored in our work as \tnshd{}, \tnres{} and \tnmod{}. We show a comparison of all the approaches in Table~\ref{tab:tn_types}.

With a \tnshd{} variant of the target network, we explicitly learn a shared set of weights $\mvec{W_{i}^{s}}$ and biases $\mvec{b_{i}^{s}}$ for some layers, and use the hypernetwork only for generating instance-specific weights $\mvec{W_{i}^{h}}$ and biases $\mvec{b_{i}^{h}}$ for the remaining layers. Target networks' outputs are then computed using either shared or instance-specific set of weights, depending on the layer.

In the \tnres{} and \tnmod{} variants of the target network, for each layer $i$ we explicitly learn shared weights $\mvec{W_{i}^{s}}$ and biases $\mvec{b_{i}^{s}}$. The hypernetwork learns to generate separate, instance-specific, weights $\mvec{W_{i}^{h}}$ and biases $\mvec{b_{i}^{h}}$. Both sets of weights and biases are then either added or multiplied and the results are used to compute layer output.

\subsection{Hypernetwork weight initialization}
Hypernetworks are very sensitive to the learning rate and require careful weight initialization to ensure stable learning. Xavier~\cite{glorot2010understanding} or Kaiming~\cite{he2015delving} weight initializations typically used for deep neural networks are insufficient for hypernetworks, as we want to achieve the corresponding initial distribution of generated weights, but we can affect it only through the weights of the hypernetwork. To solve this issue, \cite{chang2019principled} propose hypernetwork weight initialization scheme which leads to the initial target network weights following the distribution given by either Xavier or Kaiming initialization. We adopt this process for our FMLP network, but find it insufficient for SIREN networks, as they require specific initialization described by ~\cite{sitzmann2020implicit}. Therefore, to simulate SIREN initialization in the initial stage of training, we set the starting weights for the last layer of the hypernetwork according to uniform distribution with boundaries close to zero. and set the distribution of biases to the one proposed in the SIREN paper. This makes the frequencies in the subsequent target network layers grow slowly in the initial stages of network training, which is crucial to stable learning with periodic activation functions.

\section{Experiments}

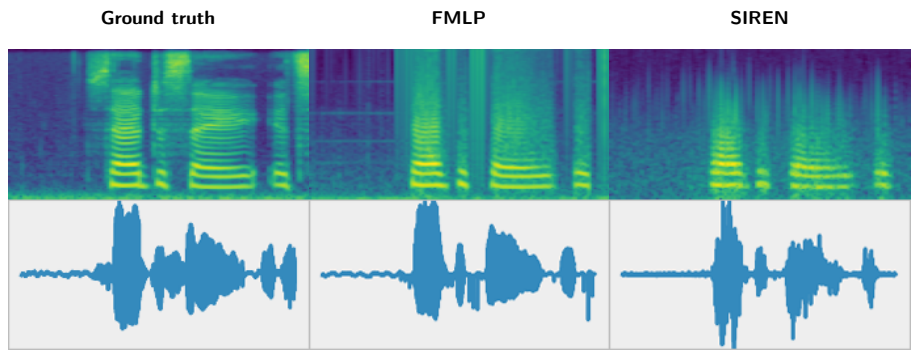
\begin{figure*}[!t]
    \begin{center}
    \begin{tikzpicture}[%
        scale=1,
        node distance=0.5cm,
        on grid,
        inner frame sep=-0.01cm,
    ]
        \clip (0, 0) rectangle (\textwidth, 4.9);
        \sffamily
        \node [inner sep=0cm, anchor=south west] at (0, 0) {\includegraphics[width=4cm,trim={0 0 0 0},clip]{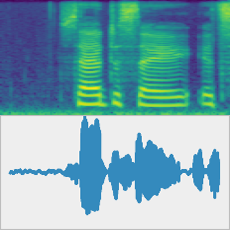}};
        \node [inner sep=0cm, anchor=south west] at (4, 0) {\includegraphics[width=4cm,trim={0 0 0 0},clip]{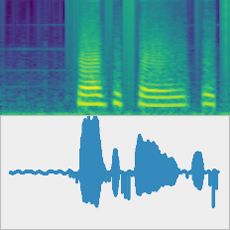}};
        \node [inner sep=0cm, anchor=south west] at (8, 0) {\includegraphics[width=4cm,trim={0 0 0 0},clip]{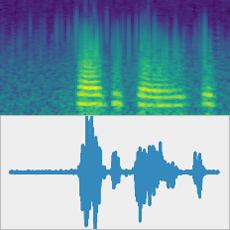}};
        \node [align=center, anchor=south] at (2, 4.225) {\scriptsize \textbf{Ground truth}};
        \node [align=center, anchor=south] at (6, 4.225) {\scriptsize \textbf{FMLP}};
        \node [align=center, anchor=south] at (10, 4.225) {\scriptsize \textbf{SIREN}};
    \end{tikzpicture}
    \end{center}
    \vspace{-1em}
    \caption{Example of VCTK validation sample reconstructed with HyperSound.}
    \label{fig:reconstructions}
\end{figure*}

\begin{table*}[t]
    \caption{HyperSound reconstruction error on the VCTK validation set compared with other methods. a) We compare our method with the state-of-the-art reconstruction model RAVE. b) We test different variants of the FMLP target network with different sizes as measured by the compression ratio (CR). c) We compare the reconstruction error obtained with SIREN target networks of size equal to FMLP with CR 1 and various values of $\omega_{i}$ and $\omega_{0}$. d) We also show the upper bound of INR reconstruction obtained by training a single model for each data point, using both FMLP and SIREN networks.}
    \label{tab:reconstruction}

    \vspace*{0.5em}

    \begin{tabularx}{\textwidth}{Xlllrrrrrr}
    
    \toprule\\[-0.9em]
    
        \textbf{Model} & & & & \textbf{MSE} & \textbf{LSD} & \textbf{SI-SNR} & \textbf{PESQ} & \textbf{STOI} & \textbf{CDPAM} \\
         \textit{Ideal metric behavior:} & & &  & $\to 0$ & $\to 0$ & $\to 100$ & $\to 4.5$ & $\to 1$ & $\to 0$ \\
    
    \midrule\\[-0.9em]
    
        a) \textit{RAVE}~\cite{caillon2021rave} & & & & 0.049 & 0.99 & -23.50 & 1.75 & 0.87 & 0.35 \\
    
    \midrule\\[-0.9em]
    
        b) \textit{HyperSound FMLP} & &  \textbf{CR} &  & & & & & & \\
        &  &    10$\times$  & & 0.013 & 1.80 & -0.25  & 1.14 & 0.71 & 0.43  \\
        &  &   4$\times$  & & 0.011 & 1.80 & 1.27   & 1.17 & 0.73 & 0.42  \\
        &  &  2$\times$  & & 0.011 & 1.73 & 2.74   & 1.21 & 0.75 & 0.38  \\
        & &  1$\times$ &  & 0.009 & 1.61 & 2.99   & 1.27 & 0.76 & 0.36  \\
        &   &  0.5$\times$ & & 0.009 & 1.53 & 4.13   & 1.34 & 0.78 & \textbf{0.30}  \\
        &  &   0.125$\times$ & & \textbf{0.007} & \textbf{1.50} & \textbf{4.48}   & \textbf{1.42} & \textbf{0.80} & \textbf{0.30} \\
    
    \midrule\\[-0.9em]
    
        c) \textit{HyperSound SIREN}   &  & \textbf{$\mvec{\omega_{0}}$} & \textbf{$\mvec{\omega_{i}}$}  & & & & & \\
   &    & 30      & 30      & 0.020 & 4.14 & -4.82  & 1.18 & 0.66 & 0.39  \\
   &    & 100     & 30      & \textbf{0.018} & 3.56 & \textbf{-3.66}  & \textbf{1.19} & 0.65 & 0.38  \\
   &    & 500     & 30      & 0.033 & 2.52 & -17.59 & 1.18 & 0.65 & \textbf{0.31}  \\
   &    & 1000    & 30      & 0.036 & 2.50 & -22.63 & \textbf{1.19} & 0.66 & 0.32  \\
   &    & 2000    & 30      & 0.037 & 2.59 & -23.93 & \textbf{1.19} & \textbf{0.67} & 0.32  \\
   &    & 2000    & 100     & 0.036 & \textbf{1.63} & -22.11 & 1.18 & \textbf{0.67} & 0.36  \\

    \midrule\\[-0.9em]

        d) \textit{Individual INRs} & &  \textbf{Model} & & & & & & &   \\ 
        \textit{(upper bound)} & & FMLP & &  0.001 & 1.39 & 17.19 & 2.31 &  0.95 & 0.23 \\
        & & SIREN & & <0.001 & 1.40 & 27.53 & 3.05 & 0.97 & 0.28 \\
    
    \bottomrule\\[-0.9em]
    \end{tabularx}
    
\end{table*}

\subsection{Setting}
\label{sec:exp:setting}
For most experiments, unless explicitly stated otherwise, we use the VCTK dataset downsampled to $f=\SI{22050}{Hz}$, with recording length set to \num{32768} samples, which is around \num{1.5} seconds. We obtain inputs by cropping the original audio to the desired length. We use the recordings of the last \num{10} speakers in the VCTK dataset as a validation set. We set batch size to for all the models to \num{16}, and train each model for \num{2500} epochs with \num{10000} training samples per epoch. We use a OneCycleLR~\cite{smith2019super} optimizer, varying learning rate depending on the target network model type, as we observe that SIREN networks require smaller learning rates to maintain stability during training.

During the evaluation, we focus on the quality of reconstructions obtained with target networks. Example of these reconstructions can be seen in Figure~\ref{fig:reconstructions}. Since there is no consensus on a~single approach for the quantitative evaluation of audio quality, we assess the reconstruction results with multiple metrics. We use quantitative metrics such as MSE, Log-Spectral Distance (LSD)~\cite{liu2022neural} and SI-SNR~\cite{luo2018tasnet}. However, since these metrics do not fully capture the perceptual quality of the audio signal, we also include metrics such as PESQ~\cite{rix2001perceptual}, STOI~\cite{taal2010short} and CDPAM~\cite{manocha2021cdpam} to approximately measure how well the reconstructed signal sounds to the human ear.

Unless explicitly mentioned otherwise, all the network hyperparameters are set as described in Section~\ref{sec:method}. For training all of the hypernetworks, we use the same data augmentations as in RAVE\cite{caillon2021rave}, i.e. random crop, phase mangle, and dequantization.

\subsection{Reconstruction quality with FMLP target network}
\label{sec:exp:fmlp}

We measure the quality of reconstructions obtained on the VCTK dataset with HyperSound using different variants of the target network. We compare the quality of reconstructions with the state-of-the-art RAVE~\cite{caillon2021rave} autoencoder retrained with sampling rate of \num{22050}~Hz. We use the same set of data augmentations for training both RAVE and HyperSound, and train RAVE for \num{3e+6} steps.

We consider multiple variants of target networks that match the desired compression ratio (measured as a ratio between the number of samples in the input recording and the number of target network parameters). We set the hypernetwork learning rate to $10^{-4}$ and use the AdamW optimizer. All target networks use five hidden dense layers with biases, and we set the width of each hidden layer to an equal value, so that we obtain the desired compression ratio. We explore more realistic scenarios with compression ratios above 1$\times$, but also include the results for target networks with higher parameter counts as an insight into the representational capacities of over-parameterized networks.

We show the results for RAVE in Table~\ref{tab:reconstruction}a, and for our method with FMLP target network Table~\ref{tab:reconstruction}b. The results show that, while over-parameterization of the target network is beneficial for the reconstruction quality, the experiments with smaller target networks still obtain passable scores across all metrics. We also notice that our method outperforms RAVE in pure reconstruction metrics across all selected target network sizes, but obtains worse scores on metrics that measure spectral distance and perceptual quality. 

\subsection{Training SIREN target networks}
\label{sec:exp:siren}

Despite \cite{benbarka2022seeing} showing a close relationship between SIREN and FMLP INRs, we observe that they behave very differently during training when compared to FMLP networks. First, we notice that without the hand-crafted hypernetwork weight initialization, as described in Section~\ref{sec:method:hn_init}, SIREN networks fail to learn any meaningful signal. Additionally, we notice that SIRENs require significantly slower learning rates to converge, and we use a learning rate of $10^{-6}$ ($\sfrac{1}{100}$ the learning rate we use in FMLP networks). We would like to use high values of $\omega_0$ and $\omega_i$, which improve the network's capability to represent high frequencies, but increasing those values makes training the network very unstable and often prevents convergence to any meaningful solution altogether.

Table~\ref{tab:reconstruction}c shows the results obtained for SIREN networks with different values of $\omega_0$ and $\omega_i$. We use network architecture corresponding to the FMLP model with a~compression ratio of \num{1}$\times$. Increasing the frequencies $\omega_{0}$ and $\omega_{i}$ slightly improves the perceptual quality of the reconstructions, but makes the training process less stable. In further experiments, we focus on FMLP networks, as they produce better results and are easier to train.

\begin{table}[t]
    \label{tab:tn_vars}
    \caption{Comparison of different methods of generating INRs with hypernetworks described in Section~\ref{sec:method:tn_types}. We notice that sharing initial layers of the target network is beneficial to reconstruction quality while making the effective size of both hypernetwork and target network smaller.}
    \vspace*{0.5em}
        \begin{tabular*}{\columnwidth}{lXr@{\hskip 0.5cm}r@{\hskip 0.5cm}r}
        \toprule
        \textbf{model type\hspace{6.2cm}} & \textbf{MSE} & \textbf{LSD} & \textbf{SI-SNR} \\
        \midrule
        \textit{\tnstd}                       & 0.009 & 1.61 & 2.99   \\
        \textit{\tnmod}            & 0.009 & 1.61 & 2.77   \\
        \textit{\tnres}                   & \textbf{0.008} & \textbf{1.52} & 2.98   \\
        \textit{first layer \tnshd}                     & \textbf{0.008} & 1.58 & \textbf{3.67}   \\
        \textit{first two layers \tnshd}                & \textbf{0.008} & 1.60 & 3.44   \\
        \bottomrule
    \end{tabular*}
\end{table}

\subsection{Upper bound of the INR reconstruction quality}
\label{sec:exp:ub}
We compute our metrics for SIREN and FMLP INRs trained in vanilla fashion to get an upper bound for reconstruction error. We select a subset of the validation data, train a model for each example minimizing the mean squared reconstruction error and report the mean metric values as an upper bound for HyperSound. We only use a subset of the validation dataset, as the mean of the metrics quickly stabilises after several recordings. We use basic variants of the FMLP and SIREN networks ($\mvec{\omega_{0}}=2000$, $\mvec{\omega_{i}}=30$), with parameters other than layer weights and biases set at fixed values as described in Section~\ref{sec:method:inr} and the basic model size as described in previous Sections.

The comparison of our results is described in Table~\ref{tab:reconstruction}d. Unsurprisingly, vanilla INRs achieve significantly better quality compared to our method and RAVE. Contrary to results for HyperSound, SIREN INRs seem to work better than FMLP models when optimized directly for a single signal. We hypothesise that this can be explained by the increased complexity of hypernetwork optimization and ReLU nonlinearity being more robust to weight perturbations.

\subsection{Exploring different ways to use weights generated by the hypernetwork}
\label{sec:exp:tn_types}

We explore possible extensions of the hypernetwork framework for INR generation, comparing three approaches described in Section~\ref{sec:method:tn_types} with the \tnstd{} variant of the target network. We conduct experiments using the network architecture corresponding to the FMLP with compression ratio \num{1}$\times$ from Section~\ref{sec:exp:fmlp}. 

The results of our experiments are described in Table~\ref{tab:tn_vars}. We notice that learning a shared set of weights for the first layer in the target network leads to better results than in the case of the \tnstd{} model, but sharing more than one layer is not beneficial. Sharing weights of the first layer can be thought of as learning a shared feature extractor for smaller hypernetwork and target network models, which means that effectively we reduce the computational cost of the method and provide more compressed representations. Other variants such as \tnmod{} and \tnres{} do not improve the results noticeably while inducing additional computational cost due to the added complexity.

\subsection{Higher-frequency weighting in the loss function}
\label{sec:exp:freqs}

We notice that our solution tends to produce representations that focus on lower frequencies, which lowers the perceptual quality of the reconstructions. We aim to improve this quality by assigning higher weights to the higher frequencies in the STFT loss, as described in Section~\ref{sec:method:optim}. As shown in Table~\ref{tab:freq}, this simple modification improves the reconstruction quality of the model, which indicates that audio processing can greatly benefit from domain-specific loss functions.

\begin{table}[t]
    \begin{minipage}[t]{0.53\textwidth}
    \caption{Impact of the frequency weighting. A higher value of $p$ indicates a stronger bias towards high frequencies in loss function, while $p=0$ equals no weighting. Our results indicate that focusing on high-frequency content in signals leads to better reconstructions.}
    \label{tab:freq}
    \vspace*{0.5em}
        \begin{tabularx}{\columnwidth}{Xr@{\hskip 0.4cm}r@{\hskip 0.4cm}r}
        \toprule
        \textbf{p} & \textbf{MSE} & \textbf{LSD} & \textbf{SI-SNR} \\
        \midrule
        \textit{0.0 [baseline]} & 0.009 & 1.61 & 2.99   \\
        \textit{0.2}            & 0.008 & 1.50 & 3.38   \\
        \textit{0.5}            & 0.008 & 1.57 & 3.09   \\
        \textit{1.0}            & \textbf{0.007} & \textbf{1.48} & \textbf{3.60}   \\
        \bottomrule
        \end{tabularx}
    \end{minipage}\hfill
    \begin{minipage}[t]{0.43\textwidth}
    \caption{Comparison of HyperSound reconstruction quality on different datasets. Our method still performs reasonably well on datasets other than VCTK, without any additional hyperparameter tuning.}
    \label{tab:datasets}
    \vspace*{0.5em}
        \begin{tabularx}{\columnwidth}{Xr@{\hskip 0.2cm}r@{\hskip 0.2cm}r}
        \toprule
        \textbf{Dataset} & \textbf{MSE} & \textbf{LSD} & \textbf{SI-SNR} \\
        \midrule
        \textit{VCTK}    & 0.009 & 1.61 & 2.99  \\
        \textit{LJ Speech}    & 0.014 & 1.65 & -1.08  \\
        \textit{LibriSpeech}    & 0.011 & 2.15 & 0.89 \\
        \bottomrule
        \end{tabularx}
    \end{minipage}
\end{table}

\subsection{Results on other datasets}
\label{sec:exp:datasets}

In our experiments, we focus on the VCTK dataset, as most publicly available audio datasets are either small, include data that does not match our sampling rate or input length requirements, or only include data from a single source or speaker. However, training hypernetworks requires large and diverse datasets and benefits from data augmentations, which are harder to apply in a reasonable way to music or environmental sound datasets. Therefore, we evaluate HyperSound reconstruction quality on additional speech datasets, i.e., LJ~Speech and LibriSpeech \textit{train-clean-360} datasets. We present the results of this evaluation in Table~\ref{tab:datasets}, showing that HyperSound still achieves reasonable results despite no additional dataset-specific hyperparameter tuning apart from learning rate reduction for LibriSpeech.

\subsection{Representational capability of INRs}
\label{sec:exp:vis}
\begin{figure}[t]
    \centering
    \includegraphics[width=0.8\columnwidth,trim={0.1cm 0.1cm 0.1cm 0.1cm},clip]{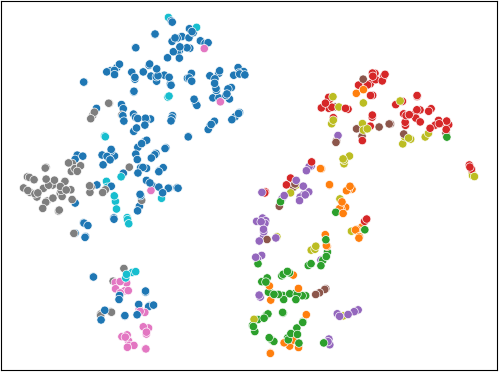}
    \vspace{-1em}
    \caption{t-SNE visualisation of the INR weights obtained on VCTK validation set. Different colors of the dots indicate different speakers. Speakers are generally clustered, and we notice a clear separation of the male and female recordings.} 
    \label{fig:tsne}
\end{figure}
Finally, we conduct a simple experiment visualizing the weights of the INRs obtained on the validation set of VCTK with t-SNE~\cite{van2008visualizing} and show the results in Figure~\ref{fig:tsne}. We group the INRs by speaker identity, as its the best-defined criterion to distinguish between recordings in the dataset. We notice that different speakers occupy distinct regions in the t-SNE subspace, with a clear separation between recordings of male and female speech. This proves that representations produced by HyperSound capture the underlying structure of the data.

\section{Conclusions}
We propose the first method of obtaining generalizable Implicit Neural Representations for audio signals, named HyperSound. We adapt the hypernetwork framework to audio processing through domain-specific modifications and demonstrate that our model performs reasonably well on the VCTK dataset, producing reconstructions quantitatively comparable to the state-of-the-art RAVE autoencoder. We investigate the use of SIREN and Fourier-mapped perceptron networks as INRs generated by the hypernetwork and try alternative ways of applying hypernetworks in the INR generation, which improve upon the standard method. INRs produced by our method can represent the signal in compressed way, using less weights than the number of samples in the original recording. Additionally, we demonstrate that our method performs reasonably across different datasets with minimal hyperparameter tuning and that the representations produced by the method align with the underlying structure of the data.

\textbf{Limitations and future work.} As theoretical guarantees of hypernetwork training are hard to define, we find our method sensitive to the hyperparameter choice and its training requires a lot of trial and error. Moreover, the misalignment between the human perception of the sound and metrics optimized in deep neural networks means that the perceptual quality of reconstructions generated with our method leaves room for improvement. We hope that further work in those directions can improve the stability of our method and better calibrate the results to human perception.

\section*{Ethical Statement}

This paper belongs to the line of fundamental research and all the experiments in this paper were performed using publicly available data. However, we acknowledge that there are potential ethical implications of our work that need to be considered.

As with all machine learning models, biases from the training data can be encoded into the model, leading to inaccurate or discriminatory behavior towards underrepresented groups.

Furthermore, the data reconstructed with Implicit Neural Representations always contains some degree of error, and models trained with different hyperparameters can produce varying representations. These properties make INRs a potential tool to evade copyright detection, and current detection algorithms are not equipped to work on data stored as INR weights, further compounding the issue.

\section*{Acknowledgements}
This work was supported by Foundation for Polish Science with Grant No POIR.04.04.00-00-14DE/18-00 carried out within the Team-Net program co-financed by the European Union under the European Regional Development Fund, and by the National Centre of Science (Poland) Grant No. 2020/39/B/ST6/01511. Filip Szatkowski and Tomasz Trzcinski are supported by National Centre of Science (Poland) Grant No. 2022/45/B/ST6/02817. Przemysław Spurek is supported by the National Centre of Science (Poland) Grant No. 2021/43/B/ST6/01456. Jacek Tabor research has been supported by a grant from the Priority Research Area DigiWorld under the Strategic Programme Excellence Initiative at Jagiellonian University.

%
%
%
\printbibliography

\end{document}